\documentclass{article}

\usepackage{microtype}
\usepackage{graphicx}
\usepackage{subcaption}
\usepackage{booktabs}
\usepackage{array}
\usepackage{hyperref}

\usepackage[accepted]{icml2026}
\usepackage{amsmath}
\usepackage{amssymb}
\usepackage{mathtools}
\usepackage{amsthm}
\usepackage[capitalize,noabbrev]{cleveref}
\crefname{appendix}{Appendix}{Appendices}
\Crefname{appendix}{Appendix}{Appendices}
\usepackage{listings}
\usepackage{xcolor}
\usepackage{algorithm}
\usepackage{algorithmic}

\theoremstyle{plain}
\newtheorem{theorem}{Theorem}
\theoremstyle{definition}

\theoremstyle{remark}

\theoremstyle{definition}
\newtheorem{example}[theorem]{Example}

\definecolor{codebg}{HTML}{F7F7F8}    
\definecolor{kwcolor}{HTML}{0F4D92}   
\definecolor{fncolor}{HTML}{6F42C1}   
\definecolor{strcolor}{HTML}{B45309}  
\definecolor{numcolor}{HTML}{2E7D32}  
\definecolor{cmtcolor}{HTML}{6A737D}  
\definecolor{opcolor}{HTML}{B91C1C}   
\definecolor{punccolor}{HTML}{8A8A8A} 

\lstdefinestyle{json}{
  basicstyle=\ttfamily\footnotesize,
  backgroundcolor=\color{codebg},
  showstringspaces=false,
  breaklines=true,
  frame=none,
  xleftmargin=4pt,
  morestring=[b]",
  stringstyle=\color{strcolor},
  morekeywords={true,false,null},
  keywordstyle=\color{numcolor}\bfseries,
  literate=
    {:}{{\textcolor{punccolor}{:}}}{1}
    {,}{{\textcolor{punccolor}{,}}}{1}
}

\lstdefinestyle{lean}{
  basicstyle=\ttfamily\small,
  backgroundcolor=\color{codebg},
  showstringspaces=false,
  breaklines=true,
  frame=none,
  xleftmargin=4pt,
  morekeywords={Input,Output,for,in,if,return,None,True,False,
                theorem,by,sorry,import,open,Type,forall,exists,fun},
  keywordstyle=\color{kwcolor}\bfseries,
  morekeywords=[2]{LLM,LeanElaborator,extract_IR,refine_IR,ir_to_lean,check},
  keywordstyle=[2]\color{fncolor},
  morestring=[b]",
  stringstyle=\color{strcolor},
  commentstyle=\color{cmtcolor}\itshape,
  literate=
    {<-}{{\textcolor{opcolor}{<-}}}{2}
}

\lstdefinestyle{leanthm}{
  basicstyle=\ttfamily\footnotesize,
  backgroundcolor=\color{codebg},
  showstringspaces=false,
  breaklines=true,
  breakatwhitespace=true,
  frame=none,
  xleftmargin=4pt,
  morekeywords={theorem,lemma,def,instance,variable,by,sorry,import,open,
                fun,let,in,if,then,else,match,with,where,Type,Prop,Sort,
                forall,Exists},
  keywordstyle=\color{kwcolor}\bfseries,
  morekeywords=[2]{Nat,Int,Real,Rat,Fin,Finset,Polynomial,Group,CommGroup,
                   Ring,CommRing,Field,Module,Submodule,Ideal,IsCompact,
                   CompactSpace,LocallyCompactSpace,Continuous,ContinuousMap,
                   MeasurableSpace,Sylow,Subgroup,Nonempty,Set,ZMod,Icc,Ico,
                   Summable,Tendsto,IsRegular,LinearEquiv,LinearMap,Not},
  keywordstyle=[2]\color{fncolor},
  morecomment=[l]{--},
  commentstyle=\color{cmtcolor}\itshape,
  literate=
    {->}{{\textcolor{opcolor}{$\to$}}}{2}
    {<->}{{\textcolor{opcolor}{$\leftrightarrow$}}}{3}
    {=>}{{\textcolor{opcolor}{$\Rightarrow$}}}{2}
    {!=}{{\textcolor{opcolor}{$\neq$}}}{2}
    {<=}{{$\leq$}}{2}
    {>=}{{$\geq$}}{2}
}

\newcommand{\TC}{TC}
\newcommand{\SF}{SF}

\newcommand{\NL}{\textsc{NL}}

\icmltitlerunning{The Signal-Coverage Matrix for Statement Autoformalization}

\begin{document}

\twocolumn[
\icmltitle{The Signal-Coverage Matrix:\\
Stratifying Type and Semantic Errors in Statement Autoformalization}

\icmlsetsymbol{equal}{*}
\begin{icmlauthorlist}
\icmlauthor{Chengxiao Dai}{usyd}
\icmlauthor{Zhaokun Yan}{tum}
\icmlauthor{Zhanhui Lin}{usyd}
\end{icmlauthorlist}
\icmlaffiliation{usyd}{School of Computer Science, University of Sydney, Sydney, Australia}
\icmlaffiliation{tum}{China Academy of Information and Communications Technology, China}
\icmlcorrespondingauthor{Chengxiao Dai}{cdai0023@uni.sydney.edu.au}
\icmlkeywords{Autoformalization, Statement formalization, Semantic faithfulness, Elaborator feedback, Signal-coverage matrix, LLM-as-judge}

\vskip 0.3in
]

\printAffiliationsAndNotice{}

\begin{abstract}
Headline type-correctness (\TC\%) of LLM autoformalization has
climbed from $\sim$53\% to $\sim$76\% in two years, yet this scalar
conceals which errors each method resolves. We propose a
signal-coverage matrix that crosses the Lean elaborator (pass/fail)
with a semantic-equivalence judgment (equivalent/not), sorting
every output into one of four cells: true success (\textsc{TS}),
type-only (\textsc{TO}), semantic-only (\textsc{SO}), or both fail
(\textsc{BF}). On ProofNet$^{\#}$ and MiniF2F-test with DeepSeek
V4-Pro across Vanilla, Lean-Retry, Sample-Filter, and Stratified
Autoformalization (SAF):
(1)~the $+34$ to $+36$ \textsc{TS} gain across the three
elab-feedback methods is $\sim$64\% type-stratum recovery, with
\textsc{SO} flat on net (87.5\% of original semantic errors rescued,
8 newly created).
(2)~The \textsc{TO}$\to$\textsc{TS} rate is $23/61$ for each method
(Wilson 95\% CI [26.6\%, 50.3\%]), and this stratum-level recovery
rate predicts $\Delta\textsc{TS}$ on held-out methods to within
$2/186$ and renders $\Delta\TC$ linear in the Vanilla elab-fail rate
across six (model, dataset) cells ($R^{2}{=}0.96$).
(3)~The two judges disagree by 26 to 37~pp on elab-feedback outputs
(vs.\ 7~pp on Vanilla), with 30 to 56\% of symbolic-judge false
negatives traceable to elaborator-forced rewrites. The persistent
residual reduces to two gold-formalization errors. \TC\% gains should be
credited by which cell moved, not by the scalar alone.
\end{abstract}

\section{Introduction}
\label{sec:intro}

Statement-level autoformalization maps a natural-language theorem to a
Lean~4 signature with a \texttt{sorry}-placeholder body. Headline
type-correctness (\TC{}\%) on ProofNet-class benchmarks
\citep{azerbayev2023proofnet,poiroux2024improving} has climbed from
$\sim$53\% (GPT-4o \citep{openai2024gpt4o}) to $\sim$76\%
(DeepSeek V4-Pro \citep{deepseek2026v4}) in two years, driven by
methods that refine against the Lean elaborator. Because \TC{}\% does
not certify denotational correctness, semantic-faithfulness (\SF{}\%)
estimators are reported alongside it: back-translation
\citep{zhou2024dtv} and generalized tree edit distance
(GTED, \citealp{liu2025gted}).

Headline metrics conceal mechanism. A $+15$~pp \TC{}\% gain hides
which error stratum is repaired and which is not, and the choice of
\SF{}\% judge can shift the reported value by $\geq 25$~pp on
elab-feedback methods (\S\ref{sec:results:judges}). No published
account decomposes \TC{}\% gains by stratum and method. We provide
such an account.

Existing methods cluster along three axes.
\textbf{Single-oracle refinement against the elaborator}
\citep{wu2022autoformalization,poiroux2024improving,atf2025,wang2024pda}
consults only the elaborator (or a close proxy) during refinement, and
the class has plateaued at $\sim$75\% \TC{}\% on ProofNet$^{\#}$
regardless of topology. \textbf{Semantic-faithfulness judging}
comprises LLM back-translation \citep{zhou2024dtv} and symbolic GTED
via the Lean LSP \citep{liu2025gted}, used as standalone \SF{}\%
estimators but never calibrated against each other at scale.
\textbf{Typed structured intermediate representations} were introduced
for Isabelle by Draft, Sketch, and Prove \citep{jiang2023dsp}, whose
sketch is a textual proof skeleton rather than a typed structural
object. The broader formal-math LM literature on premise selection,
proof generation, and math pretraining
\citep{polu2020gptf,yang2023leandojo,azerbayev2024llemma,lin2025leanstar}
addresses different stages of the pipeline and is complementary.

We instantiate a $2{\times}2$ signal-coverage matrix (elaborator
pass/fail $\times$ judge equivalent/not) on DeepSeek V4-Pro $\times$
ProofNet$^{\#}$ (186 problems) under a dual-judge protocol that pairs
Claude Opus~4.7 \citep{anthropic2026opus} with GTED at $\tau{=}0.5$,
tracking per-problem transitions from Vanilla into Lean-Retry,
Sample-Filter, and Stratified Autoformalization (SAF).

This paper makes three contributions:

(1)~\textbf{A diagnostic framework and the predictive regularity it
yields.} A $2{\times}2$ signal-coverage matrix and a $4{\times}4$
transition matrix decompose any \TC{}\% gain into type-stratum
recovery, semantic-stratum turnover, and a durable residual. Under
this lens, three algorithmically distinct methods each recover $23/61$
of Vanilla \textsc{TO} (Wilson 95\% CI [26.6\%, 50.3\%]),
and $\Delta\TC$ is linear in the Vanilla elab-fail rate at
$R^{2}{=}0.96$ across six (model, dataset)
cells (\S\ref{sec:strata},~\S\ref{sec:results:transitions},~\S\ref{sec:results:predictive}).

(2)~\textbf{A calibrated dual-judge protocol.} Opus paired with GTED
at $\tau{=}0.5$ achieves $94.4\%$ to $95.0\%$ precision across
benchmarks, and a four-class taxonomy attributes 30 to 56\% of GTED
false negatives to elaborator-forced structural rewriting
(\S\ref{sec:results:judges}).

(3)~\textbf{SAF as a causal-attribution probe.} Stratified
Autoformalization (SAF) refines a typed JSON IR against the
elaborator and emits Lean by a deterministic translator. Its
determinism makes the candidate's surface form a function of the IR
alone, attributing the $+37$~pp Opus-vs-GTED gap to elaborator-driven
surface geometry rather than LLM sampling noise
(\S\ref{sec:results:judges}). SAF ties Lean-Retry and Sample-Filter
on aggregate metrics. Its role here is diagnostic rather than
state-of-the-art optimization.

\section{The Signal-Coverage Matrix}
\label{sec:strata}

Under statement-level scope (\NL{} statement $S$ $\to$ Lean~4
signature $\sigma$ with \texttt{:= by sorry}), an output that fails
its specification falls into exactly one of two strata. A
type-stratum failure means $\sigma$ does not type-check
(\texttt{Unknown identifier}, \texttt{failed to synthesize},
\texttt{application type mismatch}, \ldots), so the math may be
understood but the Lean surface form is invalid. A semantic-stratum
failure means $\sigma$ type-checks but denotes a different
proposition from $S$, for instance hardcoding \texttt{ZMod 2} when
$S$ refers to ``any field of size~2'', using \texttt{Summable f}
when $S$ asserts the sequence has a limit, reversing an inequality,
or dropping a hypothesis. (A third, implementation stratum for
proof-level autoformalization is out of scope.)

The two refinement signals used in practice are
stratum-specialized. The Lean elaborator is complete on the type
stratum, since every failure raises an error, and silent on the
semantic stratum, since a semantically wrong but type-correct
$\sigma$ elaborates without complaint. Semantic judges, namely
back-translation \citep{zhou2024dtv} and GTED \citep{liu2025gted},
cover the semantic stratum but forgive the type stratum, since a
non-compiling fragment back-translates to noisy \NL{}.

\begin{figure}[htbp]
\centering
\includegraphics[width=\columnwidth]{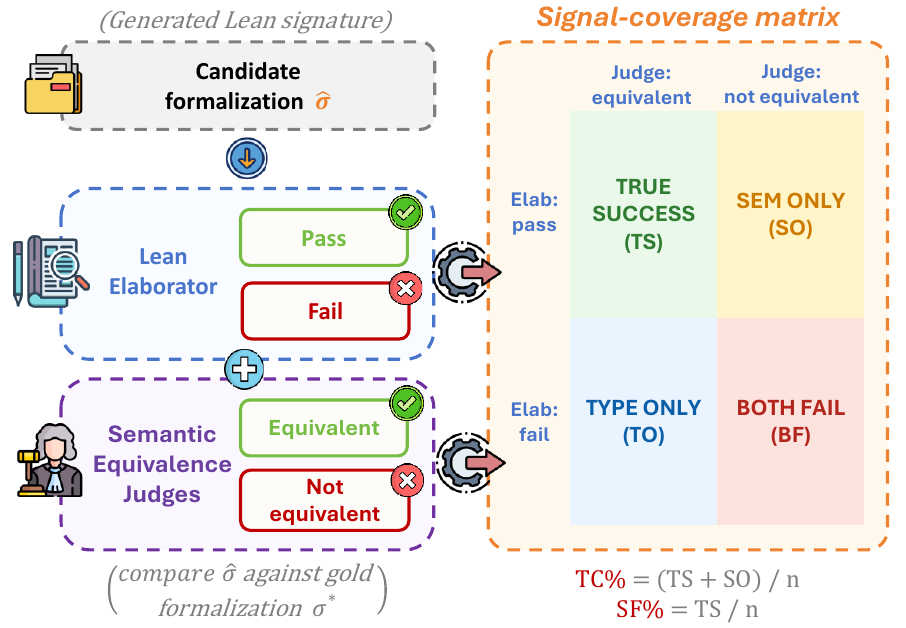}
\caption{Signal-coverage matrix: elaborator (pass/fail) $\times$ judge
(equivalent/not) gives the $2{\times}2$ cell decomposition and
\TC\%/\SF\%.}
\label{fig:eva-framework}
\end{figure}

\label{para:matrix}%
Crossing elaboration (pass / fail) with the semantic judgment
(equivalent / not-equivalent) partitions $n$
outputs into four cells (\Cref{fig:eva-framework}): true success
(\textsc{TS}), type-only failure (\textsc{TO}), semantic-only
failure (\textsc{SO}), and both fail (\textsc{BF}). These labels
denote failure modes, not correctness modes. The two off-diagonal
cells are exactly the per-instance disagreements between elaborator
and judge, and their size, sources, and dynamics across methods are
this paper's core empirical object. Writing $|c|$ for the count in
cell $c$,
\begin{equation}\label{eq:metrics}
\TC\% = \tfrac{|\textsc{TS}| + |\textsc{SO}|}{n}, \quad
\SF\% = \tfrac{|\textsc{TS}|}{n}, \quad
n = \textstyle\sum_c |c|.
\end{equation}

\section{Stratified Autoformalization}
\label{sec:saf}

Because refinement signals are stratum-specialized
(\S\ref{sec:strata}), SAF factors autoformalization through a typed
intermediate representation (IR). The LLM emits an IR, a deterministic
$\sim$30-line translator produces Lean from it, and on elab failure
the IR is refined rather than Lean source. This determinism is central:
every refined SAF output is a function of the IR alone. On the
aggregate metrics, SAF ties Lean-Retry and Sample-Filter on both
\TC{} and Opus-judged \SF{} (\S\ref{sec:results}). Its role in this
paper is the causal-attribution probe of \S\ref{sec:results:judges},
where the $+37$~pp Opus-vs-GTED gap is attributed to
elaborator-driven surface geometry rather than to LLM sampling noise.

The refinement loop (\Cref{alg:saf}) alternates IR generation by the
LLM with elaborator checking until the candidate type-checks. The
total LLM budget is $K{+}1$ (one extraction plus $K$ refinements),
and we use $K{=}3$ in all experiments.

A SAF IR is a JSON object with five fields:

\begin{lstlisting}[style=json]
{
  "imports":   ["Mathlib"],
  "opens":     ["Polynomial", "Real"],
  "objects":   [
    {"name":"G","type":"Type*", "binding":"implicit"},
    {"name":"", "type":"Group G","binding":"instance"},
    {"name":"a","type":"G",      "binding":"explicit"}
  ],
  "hypotheses":[{"id":"h", "expression":"a*b = b*a"}],
  "conclusion": "<Lean 4 expression>"
}
\end{lstlisting}

\begin{algorithm}
\caption{SAF refinement loop}
\label{alg:saf}
\begin{algorithmic}[1]
\REQUIRE NL statement $S$, retry budget $K$
\ENSURE type-checked Lean signature, or $\bot$
\STATE $\mathrm{ir} \gets \mathrm{LLM.extract\_IR}(S)$
\FOR{$k = 0$ to $K$}
    \STATE $\sigma \gets \mathrm{ir\_to\_lean}(\mathrm{ir})$
    \STATE $(\mathrm{ok}, \mathrm{errors}) \gets \mathrm{LeanElaborator.check}(\sigma)$
    \IF{$\mathrm{ok}$}
        \STATE \textbf{return} $\sigma$
    \ENDIF
    \STATE $\mathrm{ir} \gets \mathrm{LLM.refine\_IR}(S, \mathrm{ir}, \mathrm{errors})$
\ENDFOR
\STATE \textbf{return} $\bot$
\end{algorithmic}
\end{algorithm}

\noindent The \texttt{imports} and \texttt{opens} fields carry the
Mathlib environment, \texttt{objects} declares typed binders with
their binding mode, \texttt{hypotheses} lists named premises, and
\texttt{conclusion} holds the goal as a Lean~4 expression. The
\texttt{ir\_to\_lean} translator is a $\sim$30-line template
substitution that maps each field to its Lean counterpart:
\texttt{imports} to \texttt{import} lines, \texttt{opens} to
\texttt{open} lines, each \texttt{objects} entry to a typed binder
(\texttt{(x : T)} / \texttt{\{x : T\}} / \texttt{[T]} per
\texttt{binding}), \texttt{hypotheses} to hypothesis binders, and
\texttt{conclusion} to the theorem body ending in \texttt{:= by
sorry}. The translator is purely template-based (no LLM, no fuzzy
matching, no error recovery), and malformed IR fails fast.

\section{Experimental Setup}
\label{sec:setup}

\paragraph{Datasets.} ProofNet$^{\#}$ \citep{poiroux2024improving}
is the type-complex benchmark: 186 undergraduate-textbook problems
(Rudin, Munkres, Dummit-Foote, Artin, Axler, Herstein, Pugh, Putnam,
Ireland-Rosen, Shakarchi) exercising Mathlib typeclass machinery
(\texttt{Group}, \texttt{MeasurableSpace}, \texttt{ContinuousMap},
\texttt{Sylow}, \ldots). MiniF2F-test \citep{zheng2022minif2f} is
the type-simple benchmark: 244 olympiad problems admitting largely
primitive-type formalizations. The two bracket the type-system
complexity axis.

\paragraph{Subject models.} DeepSeek V4-Pro (49B active / 1.6T
total MoE parameters) is the primary subject, with the full battery
on ProofNet$^{\#}$ and three methods on MiniF2F. Qwen3.5-Plus~\citep{qwen2026}
(480B-class) and MiMo-v2.5-Pro~\citep{xiaomimimo2026} (Xiaomi, thinking traces disabled)
serve as cross-family direction checks. All decoding is at
\texttt{temperature=0.0} except where $N{=}4$ sampling is the method
itself. Gold-judge analysis is anchored on DeepSeek V4-Pro $\times$
ProofNet$^{\#}$, with cross-model and cross-dataset runs serving as
mechanism-direction checks at coarser (GTED-only) judge resolution.

\paragraph{Methods.} Four methods, three at the same
$\sim$4-LLM-call budget. Vanilla: one-shot. Lean-Retry
(\citealp{atf2025}-style): emit Lean signature, refine by feeding
elaborator errors back to the LLM, $K{=}3$ retries. Sample-Filter
(\citealp{poiroux2024improving}-style): sample $N{=}4$ candidates,
retain elab-pass candidates, pick first. SAF (this paper, \S\ref{sec:saf}):
typed JSON IR refined against the elaborator, translated to Lean by
a deterministic $\sim$30-line procedure, $K{=}3$.

\paragraph{Judges.} A dual-judge protocol pairs Claude Opus~4.7
(Anthropic, disjoint from all four subject families) as the gold
cross-judge with GTED \citep{liu2025gted} as the scalable symbolic
proxy. Opus compares candidate Lean against gold Lean for semantic
equivalence under a STRICT prompt when the candidate elaborates and
an EQUIV-IF-FIXED prompt when it does not, yielding 744 verdicts on
DeepSeek V4-Pro $\times$ ProofNet$^{\#}$ $\times$ \{Vanilla,
Lean-Retry, SAF, Sample-Filter\}, 244 on DeepSeek V4-Pro $\times$
MiniF2F $\times$ Vanilla, and 988 in total. GTED computes generalized
tree edit distance on the LSP-extracted typed operator tree, returns
a similarity in $[0,1]$, is thresholded at $\tau{=}0.5$, and runs on all 19
(model, dataset, method) combinations.

\paragraph{Infrastructure.} Lean~4.19.0 via \texttt{lean\_interact},
pre-built Mathlib, ATLAS-based \citep{liu2025atlas} GTED LSP backend, elab timeout 60s,
and JSONL resume on all pipelines.

\section{Experimental Results and Analysis}
\label{sec:results}

\subsection{Per-method cell decomposition}
\label{sec:results:matrix}

\begin{figure}[h]
\centering
\includegraphics[width=0.95\columnwidth]{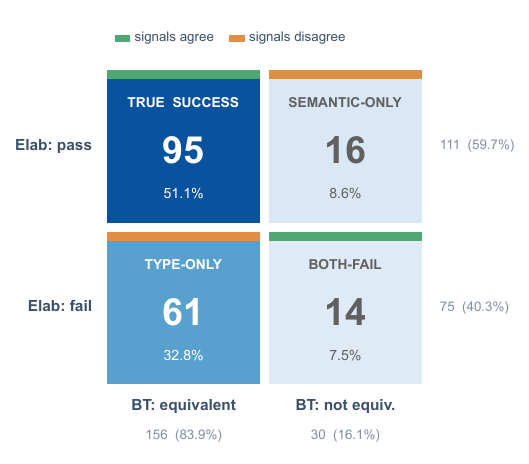}
\caption{Vanilla signal-coverage matrix (DeepSeek V4-Pro $\times$
ProofNet$^{\#}$). Rose-outlined off-diagonal cells mark
elaborator/Opus disagreements.}
\label{fig:matrix}
\end{figure}

\begin{figure}[t]
\centering
\includegraphics[width=\columnwidth]{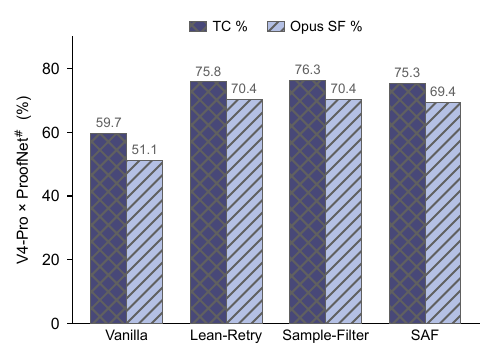}
\caption{\TC{}\% and Opus-judged \SF{}\% across four methods on
DeepSeek V4-Pro $\times$ ProofNet$^{\#}$. The three elab-feedback methods
(Lean-Retry, Sample-Filter, SAF) cluster within $\sim 1$~pp on both
axes.}
\label{fig:tcsf}
\end{figure}

For each of the 186 ProofNet$^{\#}$ test problems run on DeepSeek V4-Pro under each of
the four methods, we record the elaborator label (pass / fail) and the Opus
label (equivalent / not-equivalent), assigning each output to one
of the four cells defined in \S\ref{para:matrix}. \Cref{tab:matrix} reports the four-method matrix,
\Cref{fig:matrix} visualizes Vanilla's cell counts, and \Cref{fig:tcsf}
shows the cross-method \TC{}\,/\,\SF{} comparison (computed per
\Cref{eq:metrics}). We use DeepSeek V4-Pro $\times$ ProofNet$^{\#}$ as
the running example because it is the only (model, dataset) cell with
full four-method Opus dual-judging. \Cref{app:cross-method-dissociation}
extends the same $\Delta$\TC{}\,/\,$\Delta$\SF{} comparison to the
other five cells under GTED. The cell-level view is what lets us
separate elaborator-side recovery (\textsc{TO}$\to$\textsc{TS}) from
semantic recovery (\textsc{SO}$\to$\textsc{TS}) rather than read both
off a single aggregate rate.

\begin{table}[h]
\caption{Signal-coverage matrix on DeepSeek V4-Pro $\times$
ProofNet$^{\#}$ ($n{=}186$), four methods. Cells are defined in
\Cref{fig:eva-framework}, labels denote failure modes.}
\label{tab:matrix}
\centering
\small
\setlength{\tabcolsep}{8pt}
\begin{tabular}{@{}lrrrrrr@{}}
\toprule
\textbf{Method} & \textsc{TS} & \textsc{SO} & \textsc{TO} & \textsc{BF} & \textbf{TC\%} & \textbf{SF\%} \\
\midrule
Vanilla        &  95 & 16 & 61 & 14 & 59.7 & 51.1 \\
Lean-Retry     & \textcolor{blue}{\underline{131}} & \textcolor{blue}{\underline{10}} & 41 & \textcolor{blue}{\underline{4}} & 75.8 & \textcolor{blue}{\underline{70.4}} \\
Sample-Filter  & \textcolor{blue}{\underline{131}} & 11 & \textcolor{blue}{\underline{37}} & 7 & \textcolor{blue}{\underline{76.3}} & \textcolor{blue}{\underline{70.4}} \\
SAF            & 129 & 11 & 39 & 7 & 75.3 & 69.4 \\
\bottomrule
\end{tabular}
\end{table}

Vanilla decomposes as 95 / 16 / 61 / 14. Of the 91 failures, the
elaborator catches 75 (\textsc{TO} $+$ \textsc{BF}) and Opus catches
30 (\textsc{SO} $+$ \textsc{BF}), with overlap of 14. The 77 / 186
outputs off-diagonal (41.4\%, \textsc{SO} $+$ \textsc{TO}) are
the per-instance disagreements between the two signals. The largest
single recovery target for any elab-only feedback method is the 61
\textsc{TO} cases: semantically correct outputs that the
elaborator nonetheless rejects on surface form.

The three elab-feedback methods add 34 to 36 problems to
\textsc{TS}, shrink \textsc{TO} by 20 to 24, shrink \textsc{BF} by 7
to 10, and shrink \textsc{SO} by only 5--6 from its baseline of 16,
despite their algorithmic differences (feedback vs.\ sampling, raw
Lean vs.\ typed IR). A bootstrap 95\% CI ($B{=}10{,}000$,
\Cref{tab:bootstrap}) shows $\Delta$\textsc{TS} and
$\Delta$\textsc{TO} significant for all three methods (CI does not
cross 0), while $\Delta$\textsc{SO} is not significant for any.

\subsection{Per-problem transitions: \textsc{TS} gains come mostly from \textsc{TO}-rescue}
\label{sec:results:transitions}

\begin{figure}[t]
\centering
\includegraphics[width=0.8\columnwidth]{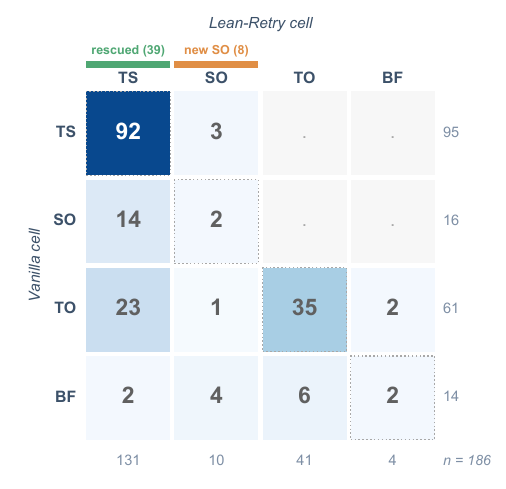}
\caption{Per-problem Vanilla~$\to$~Lean-Retry transition matrix.
Green column: rescue into \textsc{TS}. Orange column: newly
created \textsc{SO}. Dotted diagonal: stay-put. Shade encodes count.}
\label{fig:transition}
\end{figure}

Aggregate $\Delta$s answer which cells moved on net. To resolve which
problems moved, we join each problem's Vanilla cell to its Lean-Retry
cell, giving the $4{\times}4$ transition matrix in \Cref{fig:transition}
and the per-cell decomposition of $\Delta|\textsc{TS}|$ in \Cref{eq:ts-decomp}.

Writing $|c{\to}c'|$ for the number of problems in Vanilla cell $c$ that
land in Lean-Retry cell $c'$, the net \textsc{TS} gain is given by
\begin{equation}\label{eq:ts-decomp}
\begin{aligned}
\Delta|\textsc{TS}| &= \underbrace{|\textsc{TO}{\to}\textsc{TS}|}_{23}
                     + \underbrace{|\textsc{SO}{\to}\textsc{TS}|}_{14}
                     + \underbrace{|\textsc{BF}{\to}\textsc{TS}|}_{2} \\
                    &\quad - \underbrace{|\textsc{TS}{\to}\neg\textsc{TS}|}_{3}
                     \;=\; +36,
\end{aligned}
\end{equation}
attributing $23/36 \approx 64\%$ to type-stratum recovery and
$14/36 \approx 39\%$ to semantic flip on the rescued side. Per-cell
individual rescue rates are: 96.8\% of Vanilla \textsc{TS} stay
\textsc{TS}, 87.5\% (14 / 16) of Vanilla \textsc{SO}
problems are individually rescued, and only 37.7\% (23 / 61) of
Vanilla \textsc{TO} are recovered, with the remainder stuck.

\subsection{Semantic turnover and the gold-error floor}
\label{sec:results:churn}

The aggregate \textsc{SO} count drops only from 16 to 10 under
Lean-Retry, but this stability masks substantial per-problem
turnover. 14 of 16 Vanilla \textsc{SO} are individually rescued
(87.5\%), while Lean-Retry simultaneously creates 8 new \textsc{SO}
(4 from \textsc{BF}, 3 from \textsc{TS} regression, 1 from
\textsc{TO}), for a net change of $-6$: aggregate stability hides a
near-total replacement of the underlying error population.

Of the 2 Vanilla \textsc{SO} problems that remain \textsc{SO} under
Lean-Retry, both are gold-formalization bugs, verified by Opus
reasoning across all four methods. The first,
\texttt{Ireland-Rosen|exercise\_1\_30}, asks ``Prove that
$1/2 + 1/3 + \cdots + 1/n$ is not an integer.'' Gold's summand
$\frac{1}{n+2}$ does not depend on the summation index, so the gold
sum evaluates to $n/(n+2)$, a different proposition from the
harmonic tail. The candidate faithfully formalizes the harmonic
tail, and Opus correctly judges candidate $\neq$ gold under all four
methods (\Cref{ex:goldbug}).

\begin{example}[Buggy gold formalization, $n{=}1$]\label{ex:goldbug}\hfill\\
\begin{lstlisting}[style=leanthm]
-- Gold: summand `(1 : Rat) / (n+2)` is constant in i,
-- so the sum equals n/(n+2), not the harmonic tail.
theorem exercise_1_30 {n : Nat} :
  Not (Exists fun a : Int =>
    (Finset.sum (Finset.univ : Finset (Fin n))
      fun _ => (1 : Rat) / (n+2)) = a)
\end{lstlisting}
\begin{lstlisting}[style=leanthm]
-- Candidate (elab pass; Opus != gold under all 4 methods):
-- faithful formalization of the harmonic tail.
theorem exercise_1_30 {n : Nat} (hn : 2 <= n) :
  Not (Exists fun a : Int =>
    (Finset.sum (Finset.Icc 2 n)
      fun k => (1 : Rat) / k) = a)
\end{lstlisting}
\end{example}

The companion case \texttt{Dummit-Foote|exercise\_7\_2\_2} has gold
LHS $p\mid 0$, trivially true in any commutative ring, so the
biconditional collapses to a one-sided statement (\Cref{app:cases},
Case~D). Both cases are documented with Opus traces and suggested
corrections in the supplementary note.

Across all four methods, only 1 of 186 problems is durably
\textsc{SO} (\texttt{Ireland-Rosen|exercise\_1\_30}). Under
Lean-Retry the count rises to 2, both gold bugs. Excluding the gold
bugs, zero of 186 problems are durably semantic-only across the
three elab-feedback methods, so the \textsc{SO} plateau approximates
the gold-error floor of the benchmark.

\subsection{A predictive regularity: stratum-rates are method-invariant}
\label{sec:results:predictive}

The aggregate gains of \S\ref{sec:results:matrix} and the transition
matrix of \S\ref{sec:results:transitions} naturally raise the question:
do the three elab-feedback methods recover the \emph{same} problems at
the \emph{same} rate, or do they merely converge in aggregate? For each
Vanilla cell $c \in \{\textsc{TS},\textsc{SO},\textsc{TO},\textsc{BF}\}$
and each elab-feedback method $m$, define the recovery rate
\begin{equation}
\rho_{c \to \textsc{TS}}(m) \;=\;
  \frac{|\{p : \text{Vanilla}(p)=c \,\wedge\, m(p)=\textsc{TS}\}|}{|\{p : \text{Vanilla}(p)=c\}|}.
\end{equation}
\Cref{tab:recovery-rates} reports the four rates for each of the three
elab-feedback methods.

\begin{table}[t]
\caption{Per-stratum recovery rates on DeepSeek V4-Pro $\times$
ProofNet$^{\#}$, dual-judged. $\rho_{c \to \textsc{TS}}(m)$ is the
share of Vanilla cell $c$ that lands in \textsc{TS} under method $m$.}
\label{tab:recovery-rates}
\centering
\small
\setlength{\tabcolsep}{3pt}
\begin{tabular}{@{}lrrrr@{}}
\toprule
Method & \textsc{TO}$\to$\textsc{TS} & \textsc{SO}$\to$\textsc{TS} & \textsc{BF}$\to$\textsc{TS} & \textsc{TS} retain \\
\midrule
Lean-Retry    & 37.7\% & \textcolor{blue}{\underline{87.5\%}} & 14.3\% & 96.8\% \\
SAF           & 37.7\% & 75.0\% & \textcolor{blue}{\underline{42.9\%}} & 92.6\% \\
Sample-Filter & 37.7\% & 68.8\% & 28.6\% & \textcolor{blue}{\underline{97.9\%}} \\
\midrule
mean (cross-method)  & 37.7\% & 77.1\% & 28.6\% & 95.8\% \\
half-spread          & $0$\,pp & $9.4$\,pp & $14.3$\,pp & $2.6$\,pp \\
\bottomrule
\end{tabular}
\end{table}

All three elab-feedback methods recover \emph{exactly} 23 of the 61
Vanilla \textsc{TO} problems, despite differing in algorithm
(sequential elaborator feedback, deterministic IR refinement,
parallel $N{=}4$ sampling), in LLM call topology, and in candidate
generation entropy. The point estimate has a wide Wilson 95\% CI of
$[26.6\%, 50.3\%]$ at $n{=}61$, so we do not claim the true rate is
precisely 37.7\%. What we claim, and what the predictive regularity of
\Cref{tab:predicted-vs-obs} rests on, is that the three methods
recover statistically indistinguishable populations on the type
stratum: per-method CIs overlap entirely, and a Fisher exact test of
pairwise homogeneity returns $p{=}1.00$ for all three pairs.
\Cref{app:recovery-ci} gives Wilson CIs for every cell of
\Cref{tab:recovery-rates} and discusses the limits of the $n{=}186$
sample. The semantic recovery rate is high and tighter than the
\textsc{BF} rate (the latter is small-$n$ noise, 14 \textsc{BF}
problems total). The retention rate on \textsc{TS} is uniformly
$\geq 92.6\%$, so net regression is small.

Taking any single elab-feedback method's rates and applying them as
a mass-balance rule to the Vanilla cell distribution
$(|\textsc{TS}|,|\textsc{SO}|,|\textsc{TO}|,|\textsc{BF}|) =
(95,16,61,14)$ yields a testable prediction for the other methods. With Lean-Retry's rates,
$\hat{\Delta\textsc{TS}} = 0.377{\cdot}61 + 0.875{\cdot}16 + 0.143{\cdot}14 - (1-0.968){\cdot}95 = +36.0$.
\Cref{tab:predicted-vs-obs} compares predicted to observed: the
calibration on Lean-Retry predicts the other two methods'
$\Delta\textsc{TS}$ to within 2 of 186 problems, an order of
magnitude better than the bootstrap noise band of any cell.

\begin{table}[t]
\caption{Predicted vs.\ observed $\Delta\textsc{TS}$ on
DeepSeek V4-Pro $\times$ ProofNet$^{\#}$, using stratum-rates
calibrated from Lean-Retry.}
\label{tab:predicted-vs-obs}
\centering
\small
\begin{tabular}{@{}lrrr@{}}
\toprule
Method & Predicted $\Delta\textsc{TS}$ & Observed $\Delta\textsc{TS}$ & Residual \\
\midrule
Lean-Retry    & $+36.0$ & $+36$ & $\;\;0.0$ \\
SAF           & $+36.0$ & $+34$ & $-2.0$ \\
Sample-Filter & $+36.0$ & $+36$ & $\;\;0.0$ \\
\bottomrule
\end{tabular}
\end{table}

Extending across the six (model, dataset) cells of the cross-model
panel, where the test is necessarily at the TC level (Opus
dual-judging is anchored on the V4-Pro $\times$ ProofNet$^{\#}$
cell), the recovery rate of Vanilla elab-fails under Lean-Retry fits
a single linear regression in the Vanilla elab-fail rate $V_{\rm fail}$:
\begin{equation}\label{eq:dtc-fit}
\widehat{\Delta\TC} \;=\; 0.325 \cdot V_{\rm fail} \;+\; 2.88 \quad (R^{2} = 0.957),
\end{equation}
with per-cell residuals in $[-1.0,+1.3]$\,pp
(\Cref{tab:cross-cell-fit}). The recovery rate of elab-fails
($\Delta\TC / V_{\rm fail}$) is $41.2\%$ on ProofNet$^{\#}$ (range
$[38.3, 45.3]$) and $61.8\%$ on MiniF2F (range $[48.4, 73.3]$): the
same elab-signal recovers a strictly higher share of failures on
type-simple benchmarks because type-simple failures concentrate in
the easy surface-form band that $K{=}3$ refinement resolves.

\begin{table}[t]
\caption{Cross-cell fit of \Cref{eq:dtc-fit} on six (model, dataset)
cells under Lean-Retry. Rows are grouped by benchmark.}
\label{tab:cross-cell-fit}
\centering
\small
\setlength{\tabcolsep}{4pt}
\begin{tabular}{@{}lrrrr@{}}
\toprule
Model & $V_{\rm fail}\%$ & $\Delta\TC$ obs.\ & $\Delta\TC$ pred.\ & residual \\
\midrule
\multicolumn{5}{c}{\textit{ProofNet$^{\#}$\ (type-complex, $n{=}186$)}} \\
\midrule
DeepSeek V4-Pro   & $40.3$ & $+16.1$ & $+16.0$ & $+0.1$ \\
Qwen3.5-Plus      & $28.5$ & $+12.9$ & $+12.1$ & $+0.8$ \\
MiMo-v2.5-Pro     & $32.3$ & $+12.4$ & $+13.4$ & $-1.0$ \\
\midrule
\multicolumn{5}{c}{\textit{MiniF2F\ (type-simple, $n{=}244$)}} \\
\midrule
DeepSeek V4-Pro   & $12.7$ & $+\;6.1$ & $+\;7.0$ & $-0.9$ \\
Qwen3.5-Plus      & $\;\;6.1$ & $+\;4.5$ & $+\;4.9$ & $-0.4$ \\
MiMo-v2.5-Pro     & $13.5$ & $+\;8.6$ & $+\;7.3$ & $+1.3$ \\
\bottomrule
\end{tabular}
\end{table}

Two consequences for method comparison follow. First, a new method's
\emph{aggregate} $\Delta\TC$ is partially determined by its target's
Vanilla cell distribution, not only by the method's algorithmic
innovations: a method evaluated only on a type-complex benchmark
will look weaker than the same method on MiniF2F, by a
benchmark-modulation factor predicted from $V_{\rm fail}$ alone.
Second, the type-stratum recovery rate of $23/61$ is what current
elab-feedback methods deliver, and closing the gap to the elab-only
ceiling $(|\textsc{TS}|{+}|\textsc{TO}|)/n=83.9\%$ would require
recovering hard \textsc{TO} cases that survive $K{=}3$ refinement, a
population on which no current method makes progress
(\S\ref{sec:discussion}). The within-$2/186$ agreement is partly
self-consistency, since the three methods already converge in
aggregate (\Cref{tab:matrix}).

\subsection{Opus vs.\ GTED: the surface-form rewriting gap}
\label{sec:results:judges}

On the same 186 DeepSeek V4-Pro $\times$ ProofNet$^{\#}$ outputs,
Claude Opus~4.7 and GTED $\tau{=}0.5$ give different \SF\% numbers
(\Cref{tab:opus-vs-gted}). SAF's deterministic IR translator
isolates the cause.

\begin{table}[t]
\caption{Opus vs.\ GTED \SF\% on DeepSeek V4-Pro $\times$ ProofNet$^{\#}$ outputs.}
\label{tab:opus-vs-gted}
\centering
\small
\setlength{\tabcolsep}{7pt}
\begin{tabular}{@{}lrrrr@{}}
\toprule
Method & \TC\% & Opus \SF\% & GTED \SF\% & Gap \\
\midrule
Vanilla        & 59.7 & 51.1 & \textcolor{blue}{\underline{44.1}} & $+7.0$ \\
Lean-Retry     & 75.8 & \textcolor{blue}{\underline{70.4}} & \textcolor{blue}{\underline{44.1}} & $+26.3$ \\
SAF            & 75.3 & 69.4 & 32.3 & ${\color{blue}\underline{+37.1}}$ \\
Sample-Filter  & \textcolor{blue}{\underline{76.3}} & \textcolor{blue}{\underline{70.4}} & 43.5 & $+26.9$ \\
\bottomrule
\end{tabular}
\end{table}

On Vanilla the two judges agree to within $+7$~pp. On the three
elab-feedback methods they disagree by $+26$ to $+37$~pp, with Opus
reading \SF{} as rising and GTED reading it as flat or falling. The
mechanism: GTED penalizes surface-form normalizations (alternative
\texttt{open}s, \texttt{IsCompact univ}$\leftrightarrow$\texttt{CompactSpace},
$\mathrm{Fin}\,n\to\mathbb{R}$ vs.\
$\forall i:\mathbb{N},\,\mathrm{Icc}\,0\,1$ restatements) that
elab-feedback adopts to satisfy the elaborator, while Opus recognizes
these as semantically equivalent. \Cref{ex:munkres} is a representative instance
on \texttt{Munkres|exercise\_29\_4}: gold and candidate denote the same
$\Pi$ type, but GTED returns similarity~0.143 because the extracted
operator trees place an arrow node where the other places a $\forall$
binder.

\begin{example}[Surface-form gap, $n{=}1$]\label{ex:munkres}\hfill\\
\begin{lstlisting}[style=leanthm]
-- Gold
theorem exercise_29_4 :
  Not (LocallyCompactSpace
        (Nat -> Set.Icc (0 : Real) 1))
\end{lstlisting}
\begin{lstlisting}[style=leanthm]
-- Vanilla candidate (elab pass; Opus equiv; GTED sim 0.143)
theorem exercise_29_4 :
  Not (LocallyCompactSpace
        (forall i : Nat, Set.Icc (0 : Real) 1))
\end{lstlisting}
\end{example}

\noindent The gap is largest for SAF ($+37$~pp): with surface form
fixed by the deterministic translator, divergence is the IR's
choice, not LLM sampling noise. Additional cases
(\textsc{TO}$\to$\textsc{TS} recoveries, gold bugs) are catalogued in
\Cref{app:cases}.

\begin{table}[t]
\caption{GTED $\tau$ calibration against Opus.
$P = P(\text{Opus equiv} \mid \text{GTED} \geq \tau)$,
$R$ is recall on the Opus-equiv elab-pass set. Recommended row
$\tau{=}0.5$ in blue underline. Per-method ProofNet$^{\#}$ precision in
\Cref{tab:gted-cal-full}.}
\label{tab:gted-cal}
\centering
\small
\renewcommand{\arraystretch}{1.15}
\setlength{\tabcolsep}{11pt}
\begin{tabular}{@{}ccccc@{}}
\toprule
 & \multicolumn{2}{c}{\textit{MiniF2F, Vanilla}} & \multicolumn{2}{c}{\textit{ProofNet$^{\#}$, 4-method pool}} \\
\cmidrule(lr){2-3} \cmidrule(lr){4-5}
$\tau$ & $P$ & $R$ & $P$ & $R$ \\
\midrule
$0.3$           & $88.5\%$          & $62.0\%$          & $93.1\%$          & $75.1\%$          \\
${\color{blue}\underline{0.5}}$  & ${\color{blue}\underline{94.4\%}}$ & ${\color{blue}\underline{45.5\%}}$ & ${\color{blue}\underline{95.0\%}}$ & ${\color{blue}\underline{51.0\%}}$ \\
$0.7$           & $97.8\%$          & $23.5\%$          & $98.1\%$          & $21.2\%$          \\
\bottomrule
\end{tabular}
\end{table}

\Cref{tab:gted-cal} confirms that $\tau{=}0.5$ maintains 95.0\%
pooled precision on ProofNet$^{\#}$ and $\geq 93\%$ on every method
individually (per-method breakdown in \Cref{tab:gted-cal-full}). The
threshold is stable across distributions, so the gap is a structural property of
elab-feedback's surface rewriting rather than of judge calibration.

The Opus-equiv but GTED-non-equiv set ($n{=}38$ on Vanilla,
$n \in [62, 76]$ on the three elab-feedback methods) is what
drives the gap. We classify each case by the type of rewrite that
distinguishes the candidate from gold (rules and per-method audit in
\Cref{app:taxonomy}):

\begin{itemize}\itemsep0pt
\item \textbf{Notational (N).} Surface rewrites that preserve
  definitional shape: \texttt{A->B} vs.\
  \texttt{forall \_ : A, B}, \texttt{Real{\textasciicircum}n} vs.\
  \texttt{Fin n -> Real}, coercion (\texttt{ZMod 2} for the unique
  field of order 2), alternative \texttt{open}/namespace prefixes,
  or cosmetic renaming.
\item \textbf{Idiomatic (I).} Definitional aliases or Mathlib
  formulation swaps: \texttt{IsConj} vs.\ \texttt{exists g, b =
  g*a*g$^{-1}$}, \texttt{Summable} vs.\ \texttt{Tendsto} of partial
  sums, \texttt{CompactSpace} vs.\ \texttt{IsCompact Set.univ}, or
  \texttt{LinearEquiv}-via-\texttt{Nonempty} typeclass framing.
\item \textbf{Structural (S).} Binder, quantifier, or typeclass
  restructuring that changes the post-elaboration term: \texttt{let}
  vs.\ explicit hypothesis, instance binder vs.\ hypothesis binder,
  \texttt{[Field]} vs.\ \texttt{(h : Field)}, quantifier moved
  inside or outside an existential, \texttt{Iff} direction swap,
  biconditional introduced or collapsed, or \texttt{subtype} vs.\
  predicate-restricted set.
\item \textbf{Residual (Z).} Cases that mix multiple categories or
  resist single-category classification.
\end{itemize}

\begin{table}[t]
\caption{Taxonomy of Opus-vs-GTED disagreements on
DeepSeek V4-Pro $\times$ ProofNet$^{\#}$. Counts: problems with Opus
equiv $\wedge$ GTED $<0.5$ $\wedge$ elab-pass. Percentages: share of
each method's false-negative pool.}
\label{tab:taxonomy}
\centering
\small
\setlength{\tabcolsep}{3pt}
\begin{tabular}{@{}lrrrrr@{}}
\toprule
Method & N (\%) & I (\%) & S (\%) & Z (\%) & Total \\
\midrule
Vanilla        & 4 (10.5) & 6 (15.8) & \textcolor{blue}{\underline{14 (36.8)}} & \textcolor{blue}{\underline{14 (36.8)}} & 38 \\
Lean-Retry     & 11 (17.7) & 9 (14.5) & \textcolor{blue}{\underline{30 (48.4)}} & 12 (19.4) & 62 \\
SAF            & 11 (14.5) & 8 (10.5) & 23 (30.3) & \textcolor{blue}{\underline{34 (44.7)}} & 76 \\
Sample-Filter  & 3 (\phantom{0}4.8) & 6 (\phantom{0}9.7) & \textcolor{blue}{\underline{35 (56.5)}} & 18 (29.0) & 62 \\
\bottomrule
\end{tabular}
\end{table}

\Cref{tab:taxonomy} reads as follows. On Vanilla, 38 of 95 elab-pass
outputs ($40\%$) are surface-form rewrites Opus accepts but GTED does
not, distributed across all four categories. Under each elab-feedback
method the false-negative count grows by 24 to 38 problems.
For Lean-Retry and Sample-Filter the growth concentrates in
category S (16 of 24 and 21 of 24 added cases, respectively):
the elaborator pressures the LLM to restructure binders, typeclass
formulations, and quantifier orderings until the output type-checks.
SAF's growth is more diffuse: only 9 added in S (the smallest,
because SAF's deterministic \texttt{ir\_to\_lean} translator does
not freely restructure binders) but $+20$ in Z, reflecting the
multi-category rewrites the IR-mediated path tends to produce. The disagreement gap is therefore not random surface
variance but a measurement of elaborator-driven structural rewriting,
which is precisely the mechanism that recovers
\textsc{TO} into \textsc{TS}
(\S\ref{sec:results:transitions}).

\subsection{Cross-dataset: \textsc{SO} rate stable, \textsc{TO} rate scales with type-complexity}
\label{sec:results:cross-dataset}

DeepSeek V4-Pro $\times$ MiniF2F-test $\times$ Vanilla (244 problems, Opus-judged)
decomposes as \textsc{TS}=163, \textsc{SO}=24, \textsc{TO}=51, \textsc{BF}=6.
The \textsc{SO} rate is 9.8\%, essentially unchanged from 8.6\% on
type-complex ProofNet$^{\#}$. The \textsc{TO} rate, by contrast, drops from
32.8\% to 20.9\%. Dataset type-complexity loads the type stratum and not
the semantic stratum, providing supporting evidence at Vanilla-only
resolution for the mechanism of \S\ref{sec:results:transitions}. Full
MiniF2F dual-judge analysis across the four methods is future work.

\subsection{Cross-model: \TC{} gain robust to subject model}
\label{sec:results:cross-model}

\begin{figure*}[t]
\centering
\includegraphics[width=0.95\textwidth]{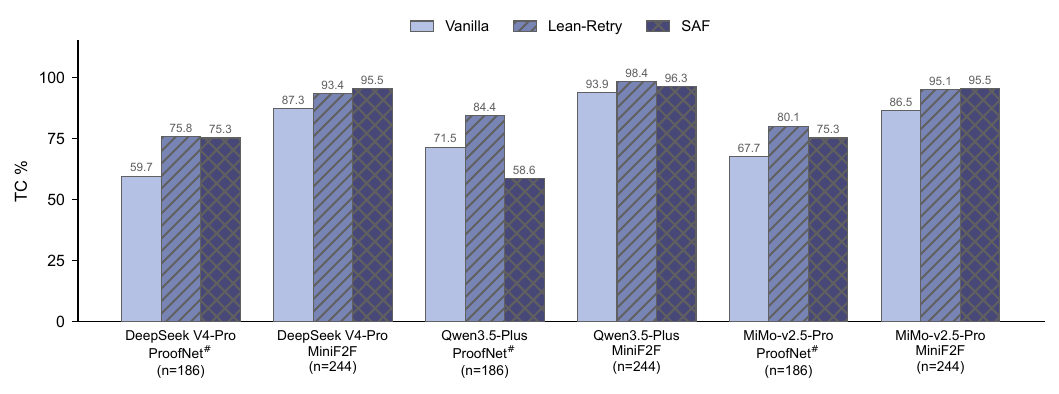}
\caption{Cross-model \TC{}\% on 3 models $\times$ 2 datasets $\times$ 3
methods.}
\label{fig:crossmodel}
\end{figure*}

\Cref{fig:crossmodel} reports \TC{}\% for three subject models on
both datasets. Lean-Retry $\geq$ Vanilla in 6/6 cells (mean
$\Delta$\TC{} $=$ $+10.1$~pp). SAF $\geq$ Vanilla in 5/6, the one
exception (Qwen $\times$ ProofNet$^{\#}$, $-12.9$~pp) being a high
IR-extraction failure rate on Qwen3.5-Plus, reported without ad-hoc patching.
Under GTED-only judging, $\Delta$\TC{} $>$ $\Delta$\SF{} holds in 6/6
cells for Lean-Retry (mean $+10.1$ vs.\ $+1.3$~pp,
\Cref{tab:cross-method-dissociation}).

\subsection{K-saturation: refinement saturates fast and three methods converge}
\label{sec:results:ksat}

\begin{figure}[h]
\centering
\includegraphics[width=\columnwidth]{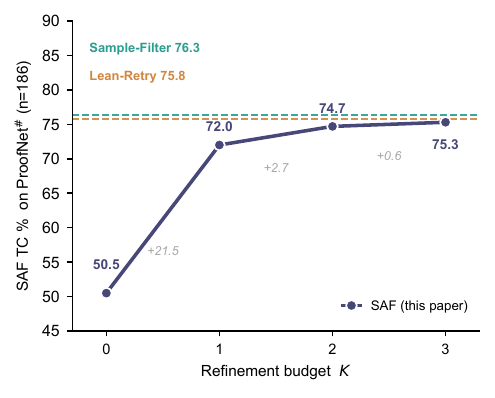}
\caption{SAF K-saturation on DeepSeek V4-Pro $\times$ ProofNet$^{\#}$.
Dashed bands: Lean-Retry and Sample-Filter saturation levels.}
\label{fig:ksat}
\end{figure}

\Cref{fig:ksat} plots SAF's K-saturation on DeepSeek V4-Pro
$\times$ ProofNet$^{\#}$. \TC{}\% rises 50.5~$\to$~72.0~$\to$~74.7~$\to$~75.3\%
across $K{=}0,1,2,3$. Refinement saturates at $K{=}1$ ($+21.5$~pp), with
$K{=}2,3$ contributing $+3.3$~pp combined. All three methods land within 1.0~pp at saturation (SAF 75.3,
Lean-Retry 75.8, Sample-Filter 76.3), so different topologies reach
the same ceiling. Per-iteration faithfulness remains high:
$k^{*}{=}0$: 89/94 (94.7\%), $k^{*}{=}1$: 34/40 (85.0\%),
$k^{*}{=}2$: 5/5 (100\%), and $k^{*}{=}3$: 1/1 (100\%). Of the 46 problems that never pass, 39 (84.8\%)
are Opus equiv-if-fixed (i.e., \textsc{TO}). The 16 durably
\textsc{TO} across all four methods share a common failure mode:
plausible but invalid Mathlib API names that elab errors cannot
prescribe (\Cref{app:cases}, Case~E).

\section{Discussion and Limitations}
\label{sec:discussion}

The (\TC{} $\neq$ \SF{}) gap itself is not new, but the
per-stratum quantification is. On type-complex benchmarks \TC{}\%
alone is misleading: the recommended reporting unit is \TC{}\%,
dual-judge \SF{}\%, and the per-cell decomposition
(\textsc{TS}/\textsc{SO}/\textsc{TO}/\textsc{BF}). Two elab-feedback
methods compared under a single symbolic judge can differ by
$\geq 25$~pp purely from structural rewriting, so independent-LLM
verification is required for any \SF{}\% claim.

\paragraph{Why three methods converge.}
The collapse to within $\sim$1~pp \TC{}\% at saturation
(\S\ref{sec:results:ksat}), the identical 37.7\%
\textsc{TO}$\to$\textsc{TS} rate
(\S\ref{sec:results:predictive}), and the shared
structural-rewriting signature in the Opus/GTED gap
(\S\ref{sec:results:judges}) are three views of the same claim: at $K{\leq}3$ the elab signal is
largely exhausted by the fixed $\sim$4-call budget, so all three
refinement topologies converge to the same family of type-checking
surface forms.
The remaining axis of progress is the signal itself, not the topology
on top of it.

Per-output semantic faithfulness is unreliable under elab-feedback
because of turnover (87.5\% rescue offset by 8 newly created errors
per method). Future signals must correlate with semantic correctness
rather than only distinguish elab-pass from elab-fail.

SAF's deterministic IR translation serves as a causal-attribution
probe for the surface-form rewriting mechanism
(\S\ref{sec:results:judges}) and the per-stratum predictive regularity
(\S\ref{sec:results:predictive}). Two ProofNet$^{\#}$ gold
formalizations contain gold-formalization errors
(\texttt{Ireland-Rosen|exercise\_1\_30},
\texttt{Dummit-Foote|exercise\_7\_2\_2}) and are documented with
suggested corrections in the supplementary note. Benchmarks aiming
to measure \textsc{SO} below 1\% will need gold audits, not
larger LLMs alone.

Our semantic labels rely on Claude Opus 4.7 as a strong cross-judge
rather than expert Lean validators ($\geq 94\%$ pooled precision in
\S\ref{sec:results:judges}). A small expert audit and a
canonicalization-aware symbolic metric (e.g.\ operator-tree matching
after \texttt{whnf} reduction) would absorb most of the N/I taxonomy
categories without an LLM judge. Both are natural follow-ups.

\section{Conclusion}
\label{sec:conclusion}

We decompose \TC{}\% gains via a $2{\times}2$ signal-coverage matrix
and a $4{\times}4$ per-problem transition matrix. On DeepSeek V4-Pro
$\times$ ProofNet$^{\#}$, three elab-feedback methods deliver +34 to
+36 \textsc{TS} via $\sim$64\% type-stratum recovery
(\Cref{eq:ts-decomp}), with
\textsc{SO} net-flat but 87.5\% per-problem
turnover. The \textsc{TO}$\to$\textsc{TS} rate is $23/61$ for each
method (Wilson 95\% CI [26.6\%, 50.3\%]), cross-method calibration
predicts $\Delta\textsc{TS}$ to within $2/186$, and $\Delta\TC{}$
fits a linear regression in the Vanilla elab-fail rate at
$R^{2}{=}0.96$ across six (model, dataset) cells. A
dual-judge protocol (Opus and GTED at 95\% pooled precision) shows
that single-judge \SF{}\% under-credits elab-feedback methods by
$\geq 25$~pp via elaborator-forced structural rewrites.

\bibliography{refs}
\bibliographystyle{icml2026}

\newpage
\onecolumn
\appendix
\crefalias{section}{appendix}

\section{Wilson 95\% CI on per-stratum recovery rates}
\label{app:recovery-ci}

\Cref{tab:recovery-ci} gives Wilson 95\% CIs for every entry of
\Cref{tab:recovery-rates}. Pairwise Fisher exact tests for homogeneity
of the \textsc{TO}$\to$\textsc{TS} rate return $p{=}1.0$ for all
three pairs (the observed counts are identical, $23/61$ for each
method). The semantic and \textsc{BF} recovery rates have overlapping
CIs across methods, consistent with a shared underlying rate plus
small-sample noise. The wide \textsc{BF} CI ($\pm{\sim}20$\,pp)
reflects $n{=}14$ rather than instability of the recovery mechanism.

\begin{table}[h]
\caption{Wilson 95\% CI for each entry of \Cref{tab:recovery-rates}.}
\label{tab:recovery-ci}
\centering
\small
\setlength{\tabcolsep}{3pt}
\begin{tabular}{@{}lllll@{}}
\toprule
Method & TO$\to$TS & SO$\to$TS & BF$\to$TS & TS retain \\
\midrule
Lean-Retry    & 37.7\% [26.6, 50.3] & 87.5\% [64.0, 96.5] & 14.3\% [\phantom{0}4.0, 39.9] & 96.8\% [91.1, 98.9] \\
SAF           & 37.7\% [26.6, 50.3] & 75.0\% [50.5, 89.8] & 42.9\% [21.4, 67.4] & 92.6\% [85.6, 96.4] \\
Sample-Filter & 37.7\% [26.6, 50.3] & 68.8\% [44.4, 85.8] & 28.6\% [11.7, 54.6] & 97.9\% [92.6, 99.4] \\
\bottomrule
\end{tabular}
\end{table}

\section{Bootstrap 95\% CI on stratum-shift $\Delta$s}
\label{app:bootstrap}

Resampling is at the problem level rather than the cell level because
the four cell counts on a given problem are coupled (each problem
contributes to exactly one cell per method), and intervals are the
percentile method on $B{=}10{,}000$ resamples (seed 20260512) on
DeepSeek V4-Pro $\times$ ProofNet$^{\#}$ ($n{=}186$). The interval
containment pattern is uniform across the three elab-feedback methods:
$\Delta$\textsc{TS} and $\Delta$\textsc{TO} are bounded away from 0
for all three, $\Delta$\textsc{SO} CIs all straddle 0, and
$\Delta$\textsc{BF} reaches significance only on Lean-Retry, which
gives the precise numerical backing for the qualitative claim of
\S\ref{sec:results:matrix}.

\begin{table}[h]
\caption{Bootstrap 95\% CI on $\Delta$ vs.\ Vanilla
($B{=}10{,}000$). $\Delta$\textsc{BF} significant only for Lean-Retry.}
\label{tab:bootstrap}
\centering
\small
\setlength{\tabcolsep}{3pt}
\begin{tabular}{@{}lrrrr@{}}
\toprule
Method & $\Delta$\textsc{TS} & $\Delta$\textsc{SO} & $\Delta$\textsc{TO} & $\Delta$\textsc{BF} \\
\midrule
Lean-Retry    & ${\color{blue}\underline{+36}}$ [$+25,+48$] & ${\color{blue}\underline{-6}}$ [$-15,+3$] & $-20$ [$-31,-10$] & ${\color{blue}\underline{-10}}$ [$-17,-3$] \\
SAF           & $+34$ [$+22,+47$] & $-5$ [$-14,+4$] & $-22$ [$-33,-11$] & $-7$ [$-15,0$] \\
Sample-Filter & ${\color{blue}\underline{+36}}$ [$+25,+47$] & $-5$ [$-14,+3$] & ${\color{blue}\underline{-24}}$ [$-35,-13$] & $-7$ [$-14,0$] \\
\bottomrule
\end{tabular}
\end{table}

\section{GTED $\tau$ calibration: per-method precision}
\label{app:gted-full}

\Cref{tab:gted-cal} of \S\ref{sec:results:judges} reports the
4-method pooled precision at $\tau{=}0.5$ ($95.0\%$ on
ProofNet$^{\#}$). The breakdown below shows that no single method
drags the pool: per-method precision stays at $\geq 93.4\%$ at
$\tau{=}0.5$ and rises monotonically with $\tau$, so the chosen
threshold is robust to which method's elab-pass outputs are pooled
into the calibration set.

\begin{table}[h]
\caption{GTED precision at three thresholds, per-method on
DeepSeek V4-Pro $\times$ ProofNet$^{\#}$ elab-pass subset.}
\label{tab:gted-cal-full}
\centering
\small
\setlength{\tabcolsep}{4pt}
\begin{tabular}{@{}lrrrr@{}}
\toprule
$\tau$ & Vanilla & Lean-Retry & SAF & Sample-Filter \\
\midrule
0.3 & 87.4\% & 95.2\% & \textcolor{blue}{\underline{95.7\%}} & 93.5\% \\
0.5 & 93.4\% & \textcolor{blue}{\underline{97.2\%}} & 94.6\% & 94.5\% \\
0.7 & 96.3\% & \textcolor{blue}{\underline{100.0\%}} & \textcolor{blue}{\underline{100.0\%}} & 96.8\% \\
\bottomrule
\end{tabular}
\end{table}

\section{Disagreement taxonomy: rules and audit}
\label{app:taxonomy}

\Cref{tab:taxonomy} of \S\ref{sec:results:judges} reports a four-class
classification of every Opus-equiv but GTED $<0.5$ output on
DeepSeek V4-Pro $\times$ ProofNet$^{\#}$, across four methods. Classes are
defined as follows:

\begin{description}\itemsep0pt
\item[N (Notational).] The rewrite preserves the operator-tree shape
modulo Lean's surface notation: \texttt{A -> B} $\leftrightarrow$
\texttt{forall \_ : A, B}, \texttt{Real{\textasciicircum}n}
$\leftrightarrow$ \texttt{Fin n -> Real}, alternative
\texttt{open}/namespace prefixes, coercion (\texttt{ZMod 2} as
representative for ``any field of order 2''),
\texttt{Function.Injective} $\leftrightarrow$ \texttt{Injective}, and
cosmetic variable renaming. GTED penalises these because its operator
tree is extracted pre-reduction.

\item[I (Idiomatic).] The rewrite swaps one Mathlib idiom for a
definitionally equivalent or near-equivalent one:
\texttt{IsConj} $\leftrightarrow$ \texttt{exists g, b = g*a*g$^{-1}$},
\texttt{Summable} $\leftrightarrow$ \texttt{Tendsto} of partial sums,
\texttt{CompactSpace} $\leftrightarrow$ \texttt{IsCompact Set.univ},
\texttt{LocallyCompactSpace} $\leftrightarrow$ \texttt{Nonempty}-wrapped
\texttt{LinearEquiv}, \texttt{generateFrom = sInf} $\leftrightarrow$
\texttt{generateFrom (sInter ...)}, etc.

\item[S (Structural).] The rewrite changes term structure: binder
ordering, hypothesis hoisting (\texttt{let}-binding vs.\ explicit
hypothesis vs.\ premise), typeclass formulation (\texttt{[Field K]}
vs.\ \texttt{(h : Field K)}, \texttt{[Fact p.Prime]} vs.\
\texttt{Nat.Primes}, \texttt{[GCDMonoid R]} vs.\ explicit gcd
predicate), quantifier reordering across an existential or universal,
\texttt{Iff} direction swap or biconditional collapse,
implication-vs-biconditional substitution,
subtype-vs-predicate-restricted-set, and hypothesis omission of
vacuously-true premises (e.g.\ \texttt{0 < n} when $n{=}0$ admits no
counterexample).

\item[Z (Residual).] Cases mixing multiple categories (e.g.\ both a
notational and a structural rewrite) or specific to NL ambiguity that
do not cleanly map to one class.
\end{description}

Classification rules are encoded as a regex cascade evaluated in the
order I~$\to$~N~$\to$~S~$\to$~Z, and the first match wins. Rules are tuned
on the Opus rationale text, then frozen before the per-method counts
were tabulated. The full classifier ships with the supplementary script
\texttt{analysis/disagreement\_taxonomy.py}, which also dumps the
per-case rationale string and assigned label for every entry in
\Cref{tab:taxonomy}. Limitations of the classifier: (i) the Z bucket
absorbs cases that mix two structural levels or that the rationale
describes ambiguously, and (ii) borderline I~vs.\ N decisions (e.g.\
\texttt{deriv}$^{[n]}$ vs.\ \texttt{iteratedDeriv n}) are made by the
first matching rule rather than by adjudication. The dominant-class
claim of \S\ref{sec:results:judges} (that S grows under elab-feedback)
is robust to these noise sources because S's share rises by 12 to 20\,pp
between Vanilla and the elab-feedback methods, well outside any
plausible reassignment of the Z residual.

\section{Cross-method \TC{}/\SF{} dissociation under GTED (6 (model, dataset) combos)}
\label{app:cross-method-dissociation}

This table reports the GTED-judged $\Delta$ vs.\ Vanilla on the
five (model, dataset) cells outside the Opus-judged
DeepSeek V4-Pro $\times$ ProofNet$^{\#}$ baseline of
\S\ref{sec:results:matrix}. The dissociation pattern reappears:
$\Delta$\TC{}\% is positive in 11 of 12 (method, model, dataset)
rows (up to $+16.1$~pp), while $\Delta$\SF{}\% stays within
$\pm 5$~pp in 11 of 12 rows and is non-positive in 6 of 12:
elab-feedback moves the elab axis without moving the semantic axis
on every cell tested.

\begin{table}[h]
\caption{$\Delta$ vs.\ Vanilla under GTED $\tau{=}0.5$ across the
six (model, dataset) cells. Left: ProofNet$^{\#}$. Right: MiniF2F.}
\label{tab:cross-method-dissociation}
\centering
\small
\setlength{\tabcolsep}{4pt}
\begin{subtable}[t]{0.48\textwidth}
\centering
\caption{ProofNet$^{\#}$\ (type-complex, $n{=}186$)}
\label{tab:cross-method-dissociation-pn}
\begin{tabular}{@{}llrr@{}}
\toprule
Model & Method & $\Delta$\TC{}\% & $\Delta$\SF{}\% \\
\midrule
DeepSeek V4-Pro  & Lean-Retry & ${\color{blue}\underline{+16.1}}$ & ${\color{blue}\underline{\phantom{-}0.0}}$  \\
DeepSeek V4-Pro  & SAF        & $+15.6$ & $-11.8$           \\
Qwen3.5-Plus     & Lean-Retry & ${\color{blue}\underline{+12.9}}$ & ${\color{blue}\underline{+4.3}}$            \\
Qwen3.5-Plus     & SAF        & $-12.9$ & $-5.4$            \\
MiMo-v2.5-Pro    & Lean-Retry & ${\color{blue}\underline{+12.4}}$ & ${\color{blue}\underline{+1.6}}$            \\
MiMo-v2.5-Pro    & SAF        & $+7.5$  & $-5.4$            \\
\bottomrule
\end{tabular}
\end{subtable}
\hfill
\begin{subtable}[t]{0.48\textwidth}
\centering
\caption{MiniF2F\ (type-simple, $n{=}244$)}
\label{tab:cross-method-dissociation-mf}
\begin{tabular}{@{}llrr@{}}
\toprule
Model & Method & $\Delta$\TC{}\% & $\Delta$\SF{}\% \\
\midrule
DeepSeek V4-Pro  & Lean-Retry & $+6.1$  & ${\color{blue}\underline{+0.8}}$            \\
DeepSeek V4-Pro  & SAF        & ${\color{blue}\underline{+8.2}}$  & $\phantom{-}0.0$  \\
Qwen3.5-Plus     & Lean-Retry & ${\color{blue}\underline{+4.5}}$  & $+0.8$            \\
Qwen3.5-Plus     & SAF        & $+2.4$  & ${\color{blue}\underline{+4.9}}$            \\
MiMo-v2.5-Pro    & Lean-Retry & $+8.6$  & ${\color{blue}\underline{+0.4}}$            \\
MiMo-v2.5-Pro    & SAF        & ${\color{blue}\underline{+9.0}}$  & $-3.7$            \\
\bottomrule
\end{tabular}
\end{subtable}
\end{table}

\section{Case studies}
\label{app:cases}

Each case below shows the natural-language statement, the gold
formalization, and one or more candidate outputs. Gold and candidates
are reproduced in Lean~4 syntax with implicit elaboration arguments
preserved. Outcome tags use the cell labels of
\S\ref{sec:results:matrix}. The five cases together cover the four
mechanisms identified in the body: Case~A is a
\textsc{TO}$\to$\textsc{TS} recovery (\S\ref{sec:results:transitions}),
Case~B a canonical GTED false negative driven by surface-form
rewriting (\S\ref{sec:results:judges}), Cases~C and~D the two distinct
gold-formalization bug shapes that pin the durable \textsc{SO} floor
(\S\ref{sec:results:churn}), and Case~E a representative
\textsc{TO} that survives $K{=}3$ refinement across all four methods
(\S\ref{sec:results:ksat}).

\paragraph{Case~A: \textsc{TYPE\_ONLY} $\to$ \textsc{TRUE\_SUCCESS}
under Lean-Retry (\texttt{Dummit-Foote|exercise\_11\_1\_13}).}
\textbf{NL.} ``As vector spaces over $\mathbb{Q}$,
$\mathbb{R}^n \cong \mathbb{R}$, for all $n \in \mathbb{Z}^{+}$.''

\begin{lstlisting}[style=leanthm]
-- Gold (Mathlib idiom)
theorem exercise_11_1_13 {n : Nat} (hn : 0 < n) :
  Nonempty (LinearEquiv (RingHom.id Rat)
                        (Fin n -> Real) Real)
\end{lstlisting}
\begin{lstlisting}[style=leanthm]
-- Vanilla (elab fail): `Real^n` does not resolve to a type
-- in this position (no `HPow Type Nat Type` instance)
theorem exercise_11_1_13 {n : Nat} (hn : 0 < n) :
  Nonempty (LinearEquiv (RingHom.id Rat)
                        (Real^n) Real)
\end{lstlisting}
\begin{lstlisting}[style=leanthm]
-- Lean-Retry K=1 (elab pass; matches gold idiom)
theorem exercise_11_1_13 {n : Nat} (hn : 0 < n) :
  Nonempty (LinearEquiv (RingHom.id Rat)
                        (Fin n -> Real) Real)
\end{lstlisting}

Opus judges equiv to gold under both STRICT and EQUIV-IF-FIXED prompts.
One of 23 \textsc{TO}$\to$\textsc{TS} recoveries.

\paragraph{Case~B: GTED false negative on a surface-form rewrite
(\texttt{Munkres|exercise\_29\_4}).}
\textbf{NL.} ``Show that $[0,1]^{\omega}$ is not locally compact in the
uniform topology.''

\begin{lstlisting}[style=leanthm]
-- Gold
theorem exercise_29_4 :
  Not (LocallyCompactSpace (Nat -> Set.Icc (0 : Real) 1))
\end{lstlisting}
\begin{lstlisting}[style=leanthm]
-- Vanilla (elab pass; Opus equiv; GTED sim = 0.143)
theorem exercise_29_4 :
  Not (LocallyCompactSpace
        (forall i : Nat, Set.Icc (0 : Real) 1))
\end{lstlisting}

The two function types are the same $\Pi$ type with different surface
notations (\texttt{A -> B} is itself notation for \texttt{forall \_ : A,
B}). The elaborator treats them identically. The typed operator tree
GTED extracts from each is structurally divergent (arrow node vs.\
$\forall$ binder), giving the 0.143 similarity. Canonical shape of the
$+26$ to $+37$~pp Opus-vs-GTED gap on elab-feedback outputs
(\S\ref{sec:results:judges}).

\paragraph{Case~C: Durable \textsc{SEM\_ONLY} from a buggy gold
formalization (\texttt{Ireland-Rosen|exercise\_1\_30}).}
\textbf{NL.} ``Prove that $1/2 + 1/3 + \cdots + 1/n$ is not an integer.''

\begin{lstlisting}[style=leanthm]
-- Gold: summand `(1 : Rat) / (n+2)` is constant in i,
-- so the sum equals n/(n+2), not the harmonic tail.
theorem exercise_1_30 {n : Nat} :
  Not (Exists fun a : Int =>
    (Finset.sum (Finset.univ : Finset (Fin n))
      fun _ => (1 : Rat) / (n+2)) = a)
\end{lstlisting}
\begin{lstlisting}[style=leanthm]
-- Candidate (elab pass; Opus != gold under all 4 methods):
-- faithfully formalizes the harmonic tail.
theorem exercise_1_30 {n : Nat} (hn : 2 <= n) :
  Not (Exists fun a : Int =>
    (Finset.sum (Finset.Icc 2 n)
      fun k => (1 : Rat) / k) = a)
\end{lstlisting}

The gold theorem and the natural-language statement are different
propositions: the gold sum is a rational $n/(n+2)$ rather than the
harmonic tail. The only problem durably \textsc{SO} across all four
methods on DeepSeek V4-Pro $\times$ ProofNet$^{\#}$.

\paragraph{Case~D: Durable \textsc{SEM\_ONLY} from a vacuous gold
biconditional (\texttt{Dummit-Foote|exercise\_7\_2\_2}).}
\textbf{NL.} ``Let $p(x) = a_n x^n + \cdots + a_0 \in R[x]$. Prove that
$p(x)$ is a zero divisor in $R[x]$ iff there is a nonzero $b \in R$ with
$b \cdot p(x) = 0$.''

\begin{lstlisting}[style=leanthm]
-- Gold: LHS `Dvd.dvd p 0` is trivially true (p * 0 = 0),
-- so the biconditional collapses to a one-sided Exists.
theorem exercise_7_2_2 {R : Type*} [Ring R]
    (p : Polynomial R) :
  Iff (Dvd.dvd p 0)
      (Exists fun b : R => b != 0 /\ b * p = 0)
\end{lstlisting}
\begin{lstlisting}[style=leanthm]
-- Candidate (elab pass under Lean-Retry / Sample-Filter;
-- Opus != gold): captures zero-divisor non-trivially.
theorem exercise_7_2_2 {R : Type*} [CommRing R]
    (p : Polynomial R) :
  Iff (Exists fun q : Polynomial R => q != 0 /\ p * q = 0)
      (Exists fun b : R          => b != 0 /\ b * p = 0)
\end{lstlisting}

Gold's LHS is vacuous over any commutative ring, collapsing the
biconditional to its right-hand side. Durably \textsc{SO} under
Lean-Retry and Sample-Filter. Under SAF the IR refinement loop fails to
elaborate at all (\textsc{BF}). Both gold-formalization bugs are
documented with Opus reasoning traces in the supplementary note
accompanying this submission.

\paragraph{Case~E: Durable \textsc{TO} across all four methods
(\texttt{Dummit-Foote|exercise\_3\_4\_5b}).}
\textbf{NL.} ``Prove that quotient groups of a solvable group are
solvable.''

\begin{lstlisting}[style=leanthm]
-- Gold (Mathlib idiom): the class name is `IsSolvable`.
theorem exercise_3_4_5b {G : Type*} [Group G] [IsSolvable G]
    (H : Subgroup G) [Normal H] :
  IsSolvable (G / H)
\end{lstlisting}
\begin{lstlisting}[style=leanthm]
-- Vanilla (elab fail; Opus equiv-if-fixed):
-- uses `Solvable`, but the Mathlib class is `IsSolvable`.
theorem solvable_quotient {G : Type*} [Group G] [Solvable G]
    (H : Subgroup G) [H.Normal] :
  Solvable (G / H)
\end{lstlisting}
\begin{lstlisting}[style=leanthm]
-- Lean-Retry K=3 (still elab fail; Opus equiv-if-fixed):
-- the elaborator reports `unknown identifier Solvable` but does
-- not surface the correct Mathlib name. The LLM only adjusts
-- binder syntax (`[H.Normal]` -> `[h : H.Normal]`).
theorem solvable_quotient {G : Type*} [Group G] [Solvable G]
    (H : Subgroup G) [h : H.Normal] :
  Solvable (G / H)
\end{lstlisting}

Opus judges both candidates equiv-if-fixed: the mathematical content
matches gold, only the Mathlib API name is wrong. SAF and
Sample-Filter likewise fail to elaborate at $K{=}3$ on this problem.
One of 16 problems that remain \textsc{TO} across all four methods on
DeepSeek V4-Pro $\times$ ProofNet$^{\#}$: a canonical instance of the
elab signal naming a symptom without surfacing the cure.

\end{document}